%% file: root.tex
\documentclass[letterpaper, 10 pt, conference]{ieeeconf}  

\usepackage{amsmath,amssymb}

\input{latexTools}
\usepackage{float}
\usepackage{multirow}
\usepackage{multicol}
\usepackage{color}
\usepackage{booktabs}
\usepackage[caption=false]{subfig}
\usepackage{graphicx}
\usepackage{textcomp}
\usepackage{tikz}
\usepackage{xcolor-material}
\usepackage{pgfplots}
\usepackage{balance}

\usepackage{hyperref}
\usepackage{graphicx}

\IEEEoverridecommandlockouts                              

\title{\LARGE \bf
DA4Event: towards bridging the Sim-to-Real Gap for Event Cameras using Domain Adaptation
}

\author{ Mirco Planamente$^{*,1,3}$, 
         Chiara Plizzari$^{*,1}$,
         Marco Cannici$^{*,2}$,
         Marco Ciccone$^{2}$,  \\
         Francesco Strada$^{1}$,
         Andrea Bottino$^{1}$,
         Matteo Matteucci$^{2}$,
         Barbara Caputo$^{1,3}$%
\thanks{The work was partially supported by the ERC project N. 637076 RoboExNovo. We also acknowledge that the research activity herein was carried out using the IIT HPC infrastructure.}
\thanks{$^{1}$ Mirco Planamente, Chiara Plizzari, Francesco Strada, Andrea Bottino and Barbara Caputo are with Politecnico di Torino, Italy {\tt\footnotesize [name.surname]@polito.it}}
\thanks{$^{2}$ Marco Cannici, Marco Ciccone and Matteo Matteucci are with Politecnico di Milano, Italy {\tt\footnotesize [name.surname]@polimi.it}}
\thanks{$^{3}$ Mirco Planamente and Barbara Caputo are with Istituto Italiano di Tecnologia, Italy {\tt\footnotesize [name.surname]@iit.it}}

\thanks{$^*$ The authors equally contributed to this work.}
}

\begin{document}

\maketitle

\markboth{IEEE Robotics and Automation Letters. Preprint Version. Accepted June, 2021}
{Planamente \MakeLowercase{\textit{et al.}}: DA4Event: towards bridging the Sim-to-Real Gap for Event Cameras using Domain Adaptation}  
\begin{abstract}
Event cameras are novel bio-inspired sensors, which asynchronously capture pixel-level intensity changes in the form of ``events". The innovative way they acquire data presents several advantages over standard devices, especially in poor lighting and high-speed motion conditions. However, the novelty of these sensors results in the lack of a large amount of training data capable of fully unlocking their potential.
The most common approach implemented by researchers to address this issue is to leverage \textit{simulated event data}. Yet, this approach comes with an open research question: \textit{how well simulated data generalize to real data?} To answer this, we propose to exploit, in the event-based context, recent Domain Adaptation (DA) advances in traditional computer vision, showing that DA techniques applied to event data help reduce the \textit{sim-to-real} gap. To this purpose, we propose a novel architecture, which we call {Multi-View DA4E} ({MV-DA4E}), that better exploits the peculiarities of frame-based event representations while also promoting domain invariant characteristics in features. Through extensive experiments, we prove the effectiveness of DA methods and {MV-DA4E} on N-Caltech101. Moreover, we validate their soundness in a real-world scenario through a cross-domain analysis on the popular RGB-D Object Dataset (ROD), which we extended to the event modality (RGB-E).

\end{abstract}

\input{Sections/1.Intro.tex}
\input{Sections/2.Related_works}

\input{Sections/3.Methods}
\input{Sections/4.Experiments}
\input{Sections/5.Conclusions}

\balance





\bibliographystyle{IEEEtran} 
\bibliography{IEEEabrv}



\end{document}

%% file: latexTools.tex
\usepackage[dvipsnames]{xcolor}
\usepackage{xcolor-material}
\usepackage{color, colortbl}

\usepackage{soul} 
\usepackage{xspace}
\usepackage{amsmath}

\usepackage{hyphenat}

\definecolor{Gray}{gray}{0.9}

\newcommand\AR[1]{\makebox[0pt][l]{$#1$}\kern0.5em\raisebox{1.5ex}{$\curvearrowright$}}
\newcommand\AL[1]{\makebox[0pt][l]{$#1$}\kern0.5em\raisebox{-1.5ex}{$\curvearrowleft$}}

\newcommand{\F}{$F$\@\xspace}
\newcommand{\DABlock}{\textit{DABlock}\@\xspace}
\newcommand{\G}{$G$\@\xspace}
\newcommand{\MVP}{\textit{MVP}\@\xspace}

\DeclareRobustCommand{\new}[1]{#1}



%% file: Sections/1.Intro.tex
\section{Introduction}

{Event-based} cameras, such as Dynamic Vision Sensors (DVS), are novel bio-inspired devices that operate in a  radically different way from conventional cameras. In fact, instead of capturing images at a fixed rate, in event-based cameras, each pixel \textit{asynchronously} emits an \textit{event} when it observes a local brightness change.
This paradigm shift allows cameras to operate at a very high dynamic range, high temporal resolution and low latency with minimal power consumption.
Additionally, their high pixel bandwidth makes them unaffected by motion blur, motivating their high potential in challenging robotics and computer vision scenarios, especially those involving objects moving at high speed and in poor lighting conditions.

Recently, novel learning approaches based on standard computer vision algorithms operating on event data achieved competitive results w.r.t. traditional approaches \cite{maqueda2018event,gehrig2019end}. 
However, training standard off-the-shelf deep learning algorithms requires a large amount of data, which is still limited by the novelty of neuromorphic cameras and by their high cost. A viable alternative to overcome the scarcity of data are event camera simulators \cite{rebecq2018esim}, which can generate reliable simulated event data. Nonetheless, an open research question arise from this approach: \textit{how well simulated data generalize to real data?}
This issue has been recently partially addressed in \cite{gehrig2020video} and \cite{stoffregen2020reducing} where authors proposed to reduce the \textit{sim-to-real gap} by acting on \textit{simulator parameters}, i.e., operating at the input level, during the data simulation phase.

\begin{figure}[t]
    \centering
    \includegraphics[width=\columnwidth]{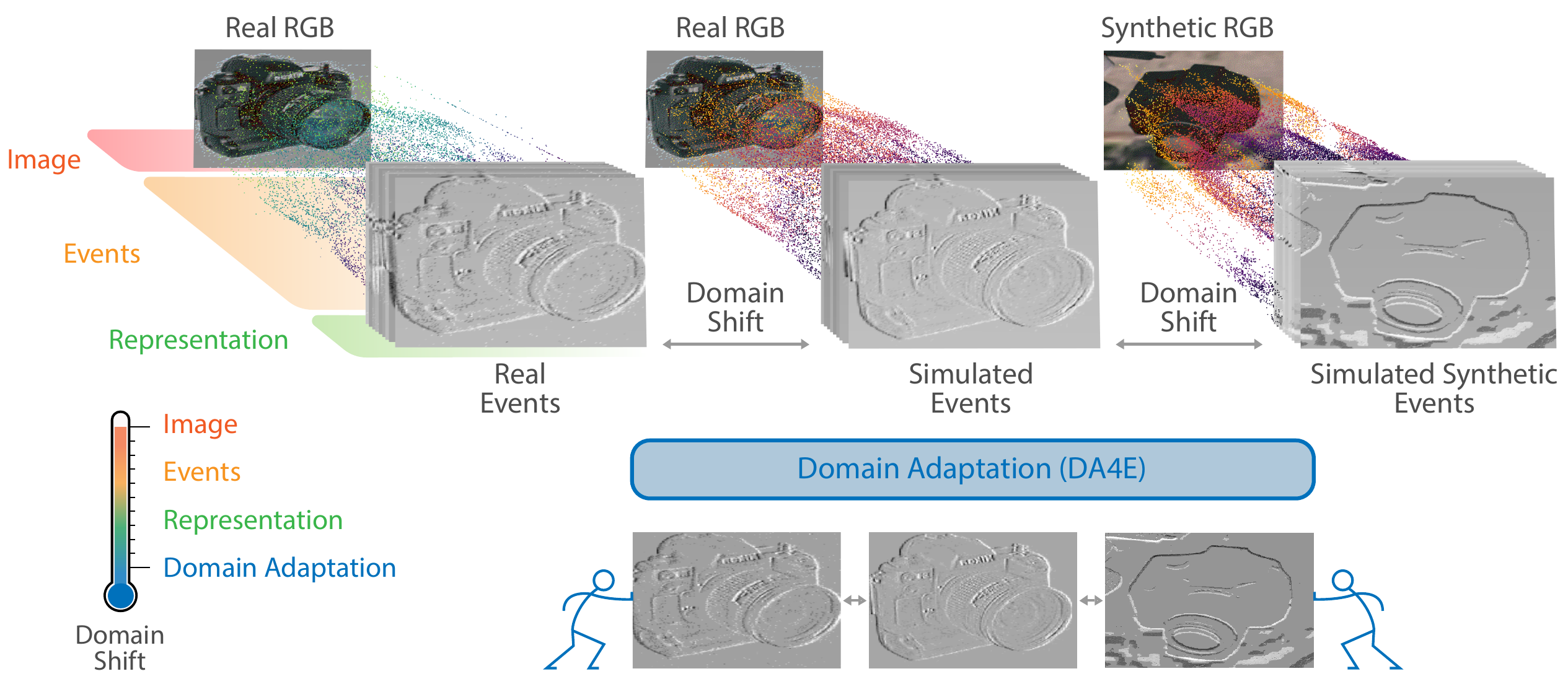}
    \caption{\textit{How can we bridge the Sim-to-Real gap in event-based cameras?} DA4Events exploits unsupervised domain adaptation techniques to solve this problem by acting at feature level. \textit{How else simulated events can be used?} We propose to use events in a real context, exploiting the complementarity with RGB data to improve networks robustness.}
    \label{fig:teaser}
\end{figure}

Our insight is that reducing this gap 
by operating \textit{at feature level}, during training, leads to more transferable representations, enhancing the generalization performance of deep networks. 
With this focus, we propose to leverage Unsupervised Domain Adaptation (UDA) techniques \cite{grl-pmlr-v37-ganin15,da-mmdlong2015learning,loghmani2020unsupervised,da-afnxu2019larger,grandvalet2005semi}, as they are specifically designed to align the distribution of features extracted from the source (simulated) and target (real) domains. 
Extensive results on the object classification task using N-Caltech101~\cite{DBLP:journals/corr/OrchardJCT15} and its simulated version Sim-N-Caltech101 \cite{gehrig2020video} prove the effectiveness of the proposed approach. In particular, we show that UDA methods are able to fill the gap between the simulated and real event domains, obtaining performance comparable to a model trained on real data. {We believe this is a significant step in unlocking event-cameras potential to new tasks, especially those requiring fine-grained annotations, as it enables to exploit the ease of simulation as well as real event sequences that are easy to gather when unlabeled.} 

Our finding unlocks novel potential uses of the event modality even when it is not possible to collect it, or the working dataset does not provide it. 
Thanks to the effectiveness of UDA methods on event-based data, we claim that RGB datasets can be extended with a \emph{simulated} event modality without hindering performance due to the sim-to-real shift (Figure \ref{fig:teaser}). This idea could be critical in many real scenarios, especially in robotics applications where simulated data are often necessary to compensate for the lack of large-scale databases. 
To prove the quality of the simulated events extracted from RGB images and their effectiveness in real-world applications, we focus on the popular RGB-D Object Dataset (ROD)~\cite{lai2011large} which provides RGB and depth modalities collected with real sensors, paired with its synthetic counterpart, SynROD~\cite{loghmani2020unsupervised}, obtained through digital rendering. We enhance both datasets with simulated events and show how the event modality, even if simulated, achieves remarkable results when used together with RGB data. 

In summary, our contributions are the following:
\begin{itemize}
    \item We propose to bridge the \textit{sim-to-real} gap for event cameras using UDA techniques, which so far are still under-explored in the event-based field, reducing the issue to a domain shift problem; 
    \item We show how the domain shift affects in different ways various event representations and to what extent different UDA approaches can soften these issues;
    \item We propose to deal with event data through a multi-view approach, called \textit{MV-DA4E}; 
    \item We extend the popular robotic dataset ROD (and its synthetic counterpart) adding event data as a novel modality, proposing a new RGB-E benchmark for object classification. 
\end{itemize}

%% file: Sections/2.Related_works.tex
\section{Related Works}
\label{sec:SOA}
    
\subsection{Unsupervised Domain Adaptation (UDA)}
The goal of UDA is reducing the domain shift between a labeled source domain and an unlabeled target one. Methods can be categorized according to their adaptation strategy. For instance, \textit{Discrepancy-based} methods explicitly minimize a distance metric among source and target distributions~\cite{da-afnxu2019larger, da-mcdsaito2018maximum, da-mmdlong2015learning}. Alternatively, \textit{adversarial-based} methods~\cite{da-adv-deng2019cluster, da-adv-tang2020discriminative} promote domain-invariant features either leveraging adversarial losses or a gradient reversal layer (GRL)~\cite{grl-pmlr-v37-ganin15}. Another possibility is to use self-supervised \textit{pretext tasks}~\cite{loghmani2020unsupervised, Xu_2019, carlucci2019domain,DBLP:journals/corr/BousmalisTSKE16}, whose losses act as an adaptation regularizer of the main loss. 
\new{Some of the above mentioned methods tackle the problem of \textit{Synth-to-Real} transfer for object classification tasks \cite{loghmani2020unsupervised,da-afnxu2019larger,grl-pmlr-v37-ganin15}, intended as the domain shift between RGB images rendered through simulation \cite{blender2018} and real RGB ones. More in general, the \textit{Synth-to-Real} transfer has been highly investigated in the semantic segmentation field. In order to address the domain shift between datasets making use of synthetic street view images \cite{ros2016synthia,richter2016playing,alberti2020idda} and real ones \cite{cordts2016cityscapes},  various \textit{adversarial-based} approaches \cite{hoffman2016fcns,tsai2018learning,vu2019advent,hoffman2018cycada} have been proposed in this field which overcome the synth-to-real shift by proposing unsupervised adversarial approaches at pixel-level.}

The research in the multi-modal field started from simple applications of existing single-modal DA methods \cite{DBLP:journals/corr/abs-1903-01212,10.1016/j.sigpro.2016.07.018}, and is now moving to more mature approaches which specifically exploit the multi-modal nature of data \cite{loghmani2020unsupervised}.

\subsection{Event Representations}
Since event cameras produce sparse encodings of the scene, these are often converted into more convenient intermediate representations before processing. Several representations have been proposed, ranging from bio-inspired ones~\cite{events-cohen2016thesis, events-cannici2019asynchronous}, to representations tailored for the subsequent processing network. Frame-like representations are by far the most widespread since they can be used together with standard computer vision algorithms and off-the-shelf convolutional neural networks (CNNs) for various tasks. A 3D volume is built by computing pixel-features resulting from accumulating events in their pixel locations. The event contribution is either computed with hand-crafted kernels~\cite{events-lagorce2016hots, events-sironi2018hats, events-zhu2019unsupervised, events-cannici2019asynchronous, events-innocenti2020temporal}, or through end-to-end trainable layers \cite{events-gehrig2019end, events-cannici2020differentiable, events-deng2020amae} able to adjust by extracting optimal features based on the task at hand. Finally, time is often discretized into bins to extract multiple representations and retain temporal resolution.

\subsection{Domain Adaptation with Events}
When training event-based deep neural models, synthetic generation of event sequences through simulation, e.g., with the ESIM~\cite{rebecq2018esim} tool, is widely used \cite{gehrig2020video,stoffregen2020reducing,rebecq2019events,DBLP:journals/corr/abs-1906-07165}. 
However, simulated events do not perfectly match data from a real sensor, introducing performance degradation. The most common mismatch source is $C$, the threshold controlling the minimum per-pixel brightness change needed to generate an event, which is typically unknown and may vary over time in a real camera. 
Gehrig~\textit{et al.} \cite{gehrig2020video} addressed this issue by randomly sampling $C$ during training, while Stoffregen \textit{et al.} \cite{stoffregen2020reducing} showed how to estimate a threshold that matches the real one, thus reducing the domain shift.
In this work, rather than acting on the generation process, we propose using UDA approaches to learn transferable representations that help improve the model's generalisation properties.

%% file: Sections/3.Methods.tex
\section{DA4Event}
\label{sec:DA2Event}

\begin{figure}
\centering
\subfloat[RGB image]{\begin{minipage}{.32\columnwidth}
    \centering
    \resizebox{\textwidth}{!}{%
    \includegraphics[width=\textwidth]{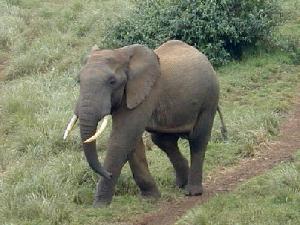}
    }
\end{minipage}}
\subfloat[Real events]{\begin{minipage}{.32\columnwidth}
    \centering
    \resizebox{\textwidth} {!} {%
    \includegraphics[width=\textwidth]{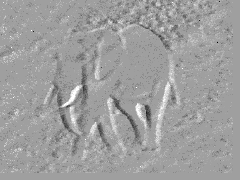}
    }
\end{minipage}}
\subfloat[Simulated events]{\begin{minipage}{.32\columnwidth}
    \centering
    \resizebox{\textwidth} {!} {%
    \includegraphics[width=\textwidth]{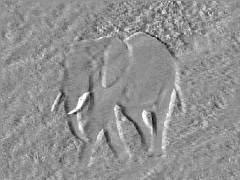}
    }
\end{minipage}}
\caption{Real and simulated events (voxel grid \cite{events-zhu2019unsupervised}) on a Caltech101 sample.}
\label{img:events_real_vs_sim}
\end{figure}

As pointed out by \cite{gehrig2020video} and \cite{stoffregen2020reducing}, the differences between simulated and real events (Figure \ref{img:events_real_vs_sim}) cause a drop in performance in several applications, independently by their representation. \new{We refer to this gap as the \textit{Sim-to-Real} gap in events.} While \cite{gehrig2020video} and \cite{stoffregen2020reducing} propose to solve the problem by acting on events' generation, our insight is to see the problem as a \textit{domain-shift} related issue. In this case, the domain shift is not in the visual appearance, as in the \new{the well-known \textit{Synth-to-Real} shift existing between rendered RGB images \cite{blender2018} and real RGB ones}. 
Indeed, the main gap is due to a different event distribution in correspondence to local brightness changes. In fact, simulators do not take into account some non-idealities typical of real cameras, such as the minimum threshold C to trigger an event, or the refractory period of event pixels, which may vary in event cameras. 

In this work, we show that by aligning the feature distribution of the source simulated domain and a target real one, UDA methods successfully reduce the \textit{Sim-to-Real} gap for event cameras, enabling neural networks to take advantage of both simulated data and real unlabeled events at training time. \new{We also extend our analysis to the \textit{Synth-to-Real} gap, by pairing both synthetic rendered images and real ones with the corresponding simulated events, showing how the simulated event modality is affected by this shift and how it can benefit from UDA techniques. This is different from a concurrent work \cite{cannici2021nrod} that studies the combined effect of \textit{Sim-to-Real} and \textit{Synth-to-Real} shifts. Indeed, \cite{cannici2021nrod} proposes an ad-hoc dataset to study this shift by providing real event data recorded through an event camera rather than obtained through simulation. However, \cite{cannici2021nrod} does not provide an analysis on the impact of the single components of this combined shift, which is instead provided in this work.  }

\subsection{Formulation}
Let us formalize the UDA problem. Our goal is to learn, on a source domain  $\mathcal{S}={\{(x^s_i,y^s_i)\}}^{N_s}_{i=1}$ with $N_s$ labeled samples associated with a (known) label space $\mathcal{Y}^s$, a model representation able to perform well on a target domain $\mathcal{T}={\{x^t_i\}}^{N_t}_{i=1}$ with $N_t$ unlabeled samples and (unknown) label space $\mathcal{Y}^t$. Our two main assumptions are that (i) the two domains have different distributions, i.e.,  $\mathcal{D}_{s} \neq \mathcal{D}_t$, and (ii) that they share the same label space, i.e., $\mathcal{Y}_{s} = \mathcal{Y}_t$. The goal is to make similar the source 
and target distributions by exploiting the UDA methods described in  Section \ref{UDA_algorithms}. 

\begin{figure*}
    \centering
    \includegraphics[width=\textwidth]{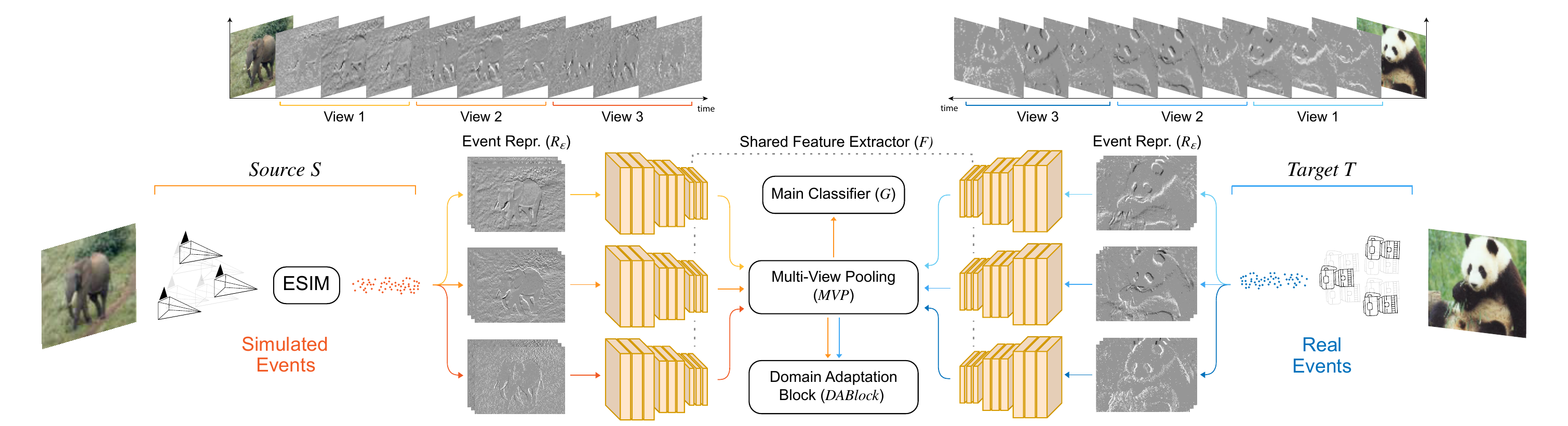}
    \caption{\new{\textbf{Top} shows the process of extracting an event representation, taking voxel grids~\cite{events-zhu2019unsupervised} and three views as an example, while \textbf{bottom} details the proposed multi-view architecture (MV-DA4E). Two unpaired random batches from \textcolor{orange}{source} and \textcolor{Cerulean}{target} domains are sampled and processed separately during training.  When the multi-view approach is not used (DA4E), event representations are fed as a single multi-channel tensor to the feature extractor $F$, and multi-view pooling is removed. Notice that only source (labelled) data are fed to the classifier $G$, while both \textcolor{Cerulean}{target} and \textcolor{orange}{source} data are fed to the DABlock.}
    }
    \label{fig:network}
\end{figure*}

\subsection{MV-DA4Event: a Multi-View Approach} \label{sec:mvda4e}
A common approach to deal with event data is to aggregate the event stream $\mathcal{E} = \{e_i = (x_i, y_i, t_i, p_i)\}_{i=1}^{N}$ describing the spatial-temporal content of the scene over a temporal period $T$, into a frame-based representation $\mathcal{R_{\mathcal{E}}} \in \mathbb{R}^{H \times W \times F}$, thus making events easily processable by off-the-shelf convolutional neural networks (CNNs). While standard RGB images encode spatial (static) information only ($R, G, B$ channels), these frame-based representations also carry temporal information, often producing a variable number of temporal channels as the event sequence is commonly split into several intervals (or bins) to retain temporal resolution, as in a video sequence. For instance, in saccadic motion, commonly used to gather event data from still planar images~\cite{DBLP:journals/corr/OrchardJCT15}, these channels correspond to the camera response to different move directions. 
As a consequence, each temporal channel represents a different observation of the recorded object, highlighting different aspects (features) of the same.

\new{A common practice in computer vision, as well as in the event-based field, is to initialize CNNs with weights pre-trained on ImageNet. However, when using a $k$-channels representation, where $k \neq 3$, the standard approach is to substitute the first convolutional block with a new one, and training it from scratch. }
This could not only limit the exploitation of the pre-trained model, but also be damaging in a cross-domain scenario. In fact, we know from the literature that the first layers of the network are usually the most affected by the domain shift~\cite{yosinski2014transferable}, thus, training them from scratch may lead the network to specialize on the source domain, poorly generalizing on the target one. Instead, when transferring pre-trained layers, the network can take advantage of robust low-level features. 


\new{Motivated by these considerations, we propose to follow a \textit{multi-view} approach to retain the first pre-trained convolutional layer.} This consists in aggregating the multi-channel event representation into three-channels images, or \textit{views},
obtaining a representation $\mathcal{\widetilde{R}_{\mathcal{E}}} \in \mathbb{R}^{H \times W \times \left \lceil F/3 \right \rceil   \times 3}$. 
A {multi-view} network (Fig. \ref{fig:network}) has been specifically designed, where each \textit{view} is fed, separately, to a feature extractor \F. The set of features thus obtained is combined with a late-fusion approach within a \MVP module, which performs average pooling, producing a $\mathbb{R}^{F_{out}}$ feature vector that is then used through the remaining parts of the network. Considering that the very first layers of the network are the more domain-specific, while the latter carry more tasks specific information, we believe that fusing the different views at the final layers of the network rather than in the earliest ones allows better generalization. 

\subsection{Network architecture}
In Figure \ref{fig:network} we outline the structure of the proposed network. Events are first obtained using the ESIM \cite{rebecq2018esim} simulator in the source domain, and directly acquired from the event-based camera in the target domain. These are split in $B$ temporal bins, and a sequence of event representations is then extracted to obtain a multi-channel volume $\mathcal{R}_\mathcal{E}$ with a multiple of $3$ channels. The representations are then collected into groups views, i.e., 3-channels frames that are treated as images and processed in parallel through a shared ResNet feature extractor \F. The set of output features is then combined in the \MVP module, which performs average pooling both spatially and across the views within features from the same domain, resulting in two feature vectors, one for each domain. Features coming from the source domain are finally used in \G to perform the final prediction, as well as in the \DABlock, together with the target ones, to perform domain adaptation. \new{Notice that during training, two completely random batches of source and target samples are selected with no match constraints between them.}








\section{UDA Algorithms} \label{UDA_algorithms}
In this section we give a brief overview of the UDA methods applied within the \DABlock of our architecture. 

\textbf{Gradient Reversal Layer (GRL).} 
The idea of GRL is to embed DA into the feature learning process. This objective is achieved by jointly optimizing the label predictor and a \textit{domain classifier} responsible for predicting whether a sample comes from the source or the target domain \cite{grl-pmlr-v37-ganin15}. Training is performed with the aim of fooling the domain classifier, maximizing its loss through a gradient reversal layer, and thus encouraging extraction of domain-invariant embeddings.



\textbf{Maximum Mean Discrepancy (MMD).} 
The method proposed by Long \textit{et al.} \cite{da-mmdlong2015learning} is based on the minimization of the \textit{Maximum Mean Discrepancy} (MMD) between source and target distributions, a metric that measures the discrepancy between them. 
%
By doing so, the final layers of the network are encouraged to produce domain-invariant features. 

\textbf{Adaptive Feature Norm (AFN).} 
Xu \textit{et al.} \cite{da-afnxu2019larger} pointed out that the main reason behind a difficult classification on target domain is due to target vectors having smaller feature norms than 
that of the source domain. 
To tackle this issue, the authors proposed to align the expectations between the $L_2$-norms of the deep embeddings of source and target domains. 
%


\textbf{Rotation (ROT).}
Xu \textit{et al.} \cite{Xu_2019} proposed to address UDA with a self-supervised task based on geometric image transformations. This auxiliary task, solved jointly with the main task, predicts the absolute image rotation of images from both the source and target domains (chosen randomly from the set $\Theta=\{0^{\circ},90^{\circ},180^{\circ},270^{\circ}\}$)
, helping the embedding model better generalize across domains. Loghmani \textit{et al.} \cite{loghmani2020unsupervised} extended ROT to multi-modal images, asking the network to predict the \textit{relative rotation} between two modalities of the same input sample, e.g., an RGB and a depth image.


\textbf{Entropy Minimization (ENT).}
Entropy minimization~\cite{grandvalet2005semi} is a widely used technique to perform UDA. It consists of representing the uncertainty on the target domain through a functional that acts as a regularization term of the classification loss. 
Adding this functional as a regularization term of the classification loss helps soften the domain shift effects between source and target distributions.  





%% file: Sections/4.Experiments.tex
\section{Experiments}
\label{sec:results}

\subsection{Datasets} 

We conduct experiments on both single- and multi-modal settings on the object classification task. We evaluate UDA for events on N-Caltech101~\cite{DBLP:journals/corr/OrchardJCT15}, and perform experiments on multimodal UDA on the RGB-D Object Dataset~\cite{lai2011large}.

\textbf{N-Caltech101.}
The Neuromorphic Caltech101 (N-Caltech101)~\cite{DBLP:journals/corr/OrchardJCT15} is an event-based conversion of the popular image dataset Caltech-101~\cite{fei2006one}. 
Samples from N-Caltech101 have been generated by recording the original RGB images using a real event-based camera moving in front of a still monitor on which still images are projected. 
An extension of N-Caltech101 has recently been proposed in \cite{gehrig2020video}, where a simulated replica of the dataset is obtained with the ESIM simulator~\cite{rebecq2018esim} by re-creating the same setup used for recording real samples. We follow \cite{gehrig2020video} and use these recordings as simulated source data and that from N-Caltech101 as target real samples.
%
\new{We use the train and test splits provided in the EST \cite{gehrig2019end} official codebase, and evaluate the proposed approach by computing the top-1 accuracy on the test set of the target real domain, as in \cite{gehrig2020video}.}

\begin{figure}
\centering
\subfloat[RGB]{\begin{minipage}{.23\columnwidth}
    \centering
    \resizebox{\textwidth}{!}{%
    \includegraphics[width=\textwidth]{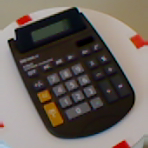}
    }
\end{minipage}}
\subfloat[Events]{\begin{minipage}{.23\columnwidth}
    \centering
    \resizebox{\textwidth} {!} {%
    \includegraphics[width=\textwidth]{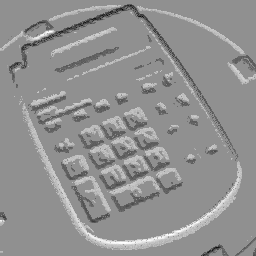}
    }
\end{minipage}}
\subfloat[Syn RGB]{\begin{minipage}{.23\columnwidth}
    \centering
    \resizebox{\textwidth} {!} {%
    \includegraphics[width=\textwidth]{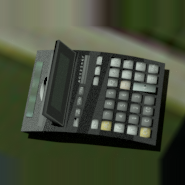}
    }
\end{minipage}}
\subfloat[Syn Events]{\begin{minipage}{.23\columnwidth}
    \centering
    \resizebox{\textwidth} {!} {%
    \includegraphics[width=\textwidth]{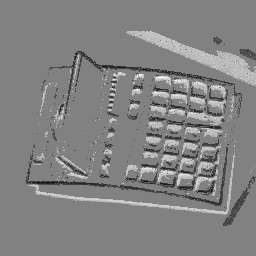}
    }
\end{minipage}}
\caption{Samples from the ROD ~\cite{lai2011large} dataset (a)-(b), and from the synthetic version synROD \cite{loghmani2020unsupervised} (c)-(d). Event sequences are displayed using a voxel-grid~\cite{events-zhu2019unsupervised} representation.}
\label{img:rod_events}
\end{figure}

\textbf{ROD.}
The RGB-D Object Dataset (ROD)~\cite{lai2011large} is one of the most common benchmarks for object recognition in robotics. It contains $41,877$ samples of $300$ items grouped in $51$ categories, captured by an RGB-D camera. 
A synthetic version of the dataset, synROD, has been recently proposed in \cite{loghmani2020unsupervised} to study the \textit{Synth-to-Real} domain-shift in multimodal settings. SynROD contains RGB-D renderings~\cite{blender2018} of 3D models from the same categories as ROD in realistic lighting conditions. We extend both versions of the dataset by introducing events as an additional third modality, following the same procedure used for N-Caltech101. 
Examples of the obtained event sequences are reported in Figure \ref{img:rod_events}. 
\new{We follow the evaluation procedure in \cite{loghmani2020unsupervised}, and report the top-1 accuracy score evaluated on real ROD samples.}

\subsection{Event-Representations}\label{ev-repr}
In this work we focus on grid-like event representations, which consist in the process of converting a stream of asynchronous events into a volume $\mathcal{R_{\mathcal{E}}} \in \mathbb{R}^{H \times W \times F}$ with $F$ features. We summarize them in the following.

\textbf{Voxel Grids.} 
This representation, also known as event volume~\cite{events-zhu2019unsupervised}, splits time in a fixed number $B$ of bins and aggregates events in their pixel locations by interpolating polarity values over time. The result is a $B$-channel representation is which each event's contribution is weighted according to its occurrence in time within the temporal bin.

\textbf{HATS.} 
The Histograms of Time Surfaces (HATS)~\cite{events-sironi2018hats} is a two channel representation combining hand-crafted features with a memory mechanism resilient to noise. The event stream is divided into a grid of non-overlapping memory cells extracting local 2D surfaces from each event's neighborhood through an exponential kernel. Surfaces from each cell are aggregated into histograms, one per polarity, and rearranged according to the position of their originating cell. Temporal resolution is lost as the entire temporal window is condensed into a single frame that does not retain temporal resolution.

\textbf{EST.} 
The Event Spike Tensor (EST)~\cite{events-gehrig2019end} is an end-to-end trainable representation. Its functioning is similar to a voxel grid, with the difference that timestamp is used as pixel feature and the kernel function used to weight events contributions is learnt by a multi-layer perceptron network. Events are grouped by polarity to extract a two-channels representation from each bin.
%
%

\textbf{MatrixLSTM.} 
MatrixLSTM~\cite{events-cannici2020differentiable} is similar to EST,  with the difference that pixel features are computed using a matrix of LSTM~\cite{hochreiter1997long} cells with shared parameters. Every cell processes the temporal-ordered sequence of events generated by each pixel and the final output of the LSTM is used as the pixel feature. The number of features can be customized and bins are optionally used to extract multiple representations.

%
%
%

\subsection{Implementation details}


We implement the proposed method within the PyTorch autodiff framework, using a ResNet34~\cite{he2016deep} as the feature extractor \F in N-Caltech101 experiments, and a ResNet18~\cite{he2016deep} in ROD ones, both pretrained on ImageNet. For a fair comparison, we use the same network configurations as in \cite{loghmani2020unsupervised} for both the object recognition classifier \G and the network used in the pretex rotation task. 
\new{We compare the proposed multi-view approach against a baseline having the same architecture, pre-trained on ImageNet, but in which event representations are directly fed as a single multi-channel tensor without view grouping. In this case, the first convolutional layer is replaced with a new randomly initialized convolution matching the number of input channels, and the multi-view pooling stage is removed.}
Event representations and RGB images going through the main backbone \F are preprocessed and augmented during training following the procedure in \cite{loghmani2020unsupervised}.
Input images are normalized with the same mean and variance used for the ImageNet pre-training, while we kept event representations un-normalized as this provided better performance. 
\new{We use $9$ bins for both voxel grids and EST representations, resulting respectively in $3$ and $6$ views, since the latter produces $2$ channels from each bin. The number of output channels can be customized in MatrixLSTM, therefore we set the layer to directly produce $3$-channel output representations and set the number of bins to $3$ as this configuration performed the best. Notice that since HATS only provides $2$ channels, without splitting the temporal frames into bins by default, we are not able to apply the proposed multi-view approach.}
We train all network configurations using SGD as optimizer, batch size $32$ and $64$ for N-Caltech101 and ROD experiments respectively, and weight decay $0.003$. We fine tune the DA losses' weights for each event representation and DA method, reporting the accuracy scores for the best configurations only, averaged over $3$ runs with different random seeds.

\subsection{Results}

\textbf{Sim $\rightarrow$ Real.}
We first assess the effectiveness of the UDA algorithms in reducing the domain-shift under the Sim-to-Real scenario using N-Caltech101. In Table \ref{tab:da_ncaltech}, we show the performance of GRL \cite{grl-pmlr-v37-ganin15}, MMD \cite{da-mmdlong2015learning}, Rotation \cite{Xu_2019}, AFN \cite{da-afnxu2019larger} and Entropy~\cite{grandvalet2005semi} against the baseline Source Only, consisting in training on labelled source data only (\textit{Sim}), and testing directly on unlabelled target data (\textit{Real}), without performing any adaptation strategy. We consider as upper-bound the performance obtained by training on real training data, and testing directly on it, in a supervised fashion (\textit{Supervised}). For each method, we report both the results obtained with (\textit{MV-DA4E}) and without (\textit{DA4E}) the proposed multi-view approach. We consider the effect of UDA strategies on two non-learnable event representations (\textit{VoxelGrid} and \textit{HATS}), and two learnable ones (\textit{EST} and \textit{MatrixLSTM}). The empirical evaluation performed allows us to answer the following research questions.
\setlength{\tabcolsep}{6.5pt}
\begin{table}[t]
\begin{center}
\vspace{3mm}
\caption{\new{Target Top-1 Test Accuracy} (\%) of UDA methods on N-Caltech101. Bold: representation's highest result. }
\label{tab:da_ncaltech}
\begin{tabular}{llcccc}
\toprule\noalign{\smallskip}
\multicolumn{6}{c}{\normalsize\textsc{N-Caltech101 (Sim $\implies$ Real) }} \\
\noalign{\smallskip}
\cline{1-6}
\noalign{\smallskip}
Method &  & \begin{tabular}[c]{@{}c@{}}Voxel\\Grid\end{tabular} & HATS & EST & \begin{tabular}[c]{@{}c@{}}Matrix\\LSTM\end{tabular}\\
\noalign{\smallskip}
\toprule\noalign{\smallskip}
\multirow{2}{*}{\centering Source Only} & \textit{baseline} &
 80.99 & 58.32 & 80.08 & 82.21 \\
& \textit{MV-baseline} & 84.59 & - & 83.07 & 84.89  \\
 
 \hline
 \noalign{\smallskip}
\multirow{2}{*}{\centering GRL \cite{grl-pmlr-v37-ganin15}} & DA4E &
 83.08 & 65.38 & 83.38 & 82.94 \\
& MV-DA4E & 86.77 & - & 84.03 & 85.75  \\

 \hline
 \noalign{\smallskip}
\multirow{2}{*}{\centering MMD \cite{da-mmdlong2015learning}} & DA4E &
 86.37 & 69.86 & 83.61 & 84.04 \\
& MV-DA4E & 88.23 & - & 85.36 & \textbf{88.05}  \\

 \hline
 \noalign{\smallskip}
\multirow{2}{*}{\centering Rotation \cite{Xu_2019}} & DA4E &
 79.13 & 61.52 & 80.69 & 83.57 \\
& MV-DA4E & 86.63 & - & 84.49 &  85.7 \\
 \hline
 \noalign{\smallskip}
\multirow{2}{*}{\centering AFN \cite{da-afnxu2019larger}} & DA4E &
 84.49 & \textbf{69.96} & 83.59 & 85.0 \\
& MV-DA4E & {88.3} & - & 85.92 & 87.59  \\
 \hline
 \noalign{\smallskip}
\multirow{2}{*}{\centering Entropy \cite{grandvalet2005semi}} & DA4E &

 87.0& 65.58& 85.54 & 85.97\\
& MV-DA4E & \textbf{89.24} & - &  \textbf{86.06} &  86.09 \\
 \hline
 \noalign{\smallskip}

\rowcolor{Gray}
& \textit{RealEvent}  & 
 88.13 & 76.45 & 88.17 & 87.65\\
\rowcolor{Gray}
\multirow{-2}{*}{Supervised } 
 & \textit{MV-RealEvent} & 90.09 & - & 89.25 & 90.35  \\
\bottomrule
\end{tabular}
\end{center}
\end{table}
\setlength{\tabcolsep}{1.4pt}

\textit{Are UDA methods useful in reducing Sim-to-Real gap?}

From the results in Table \ref{tab:da_ncaltech}, it has to be noticed that, for all event representations, in almost all the cases the UDA methods achieve better performance than the baseline Source Only, surpassing it by up to $6\%$ on VoxelGrid, $11\%$ on HATS, $6\%$ on EST and $4\%$ on MatrixLSTM. There is one single case where Rotation is on-pair with the Source Only, which is VoxelGrid without multi-view approach. A reason could be that the main benefit of Rotation is to enforce the network to focus on the geometric part of the input by solving the transformation. Indeed, event data already encodes geometric information (e.g., movement direction), thus Rotation could, in some situations, be potentially unhelpful. In fact, the network could learn to find a trivial solution (shortcut) to solve the pretext task \cite{noroozi2016unsupervised}, for instance analyzing the movement direction over the edges.

Interestingly, we can see that not all representations suffer in the same way from the domain shift. For instance, it is noticeable that HATS  is the representation suffering the most from the Sim-to-Real shift, as performance decreases by up to $16\%$ when testing directly on target domain (Source Only) rather that on source (Supervised). Intuitively, the reason is intrinsic in the representation itself. Indeed, when representing events with HATS the temporal resolution is lost (see Section \ref{ev-repr}), potentially causing a degradation in performance when testing on data belonging to a different distribution. It has to be remarked that by performing a complete \textit{DA4E} benchmark we found the optimal combination which allows us to fill the Sim-to-Real gap. 
\new{In Figure \ref{fig:comment_8} we showcase the scalability of our approach when the access to target data is limited by showing how the performance of the proposed methods changes when only a percentage of target data is available during training ($25\%,50\%,75\%$). It can be noticed that an improvement by up to $4\%$ over the  \textit{source only} baseline ($0\%$ of training target data) is guaranteed, even when a very small percentage of target samples is available.}

\begin{figure}[t]
    \centering
    \includegraphics[width=\columnwidth]{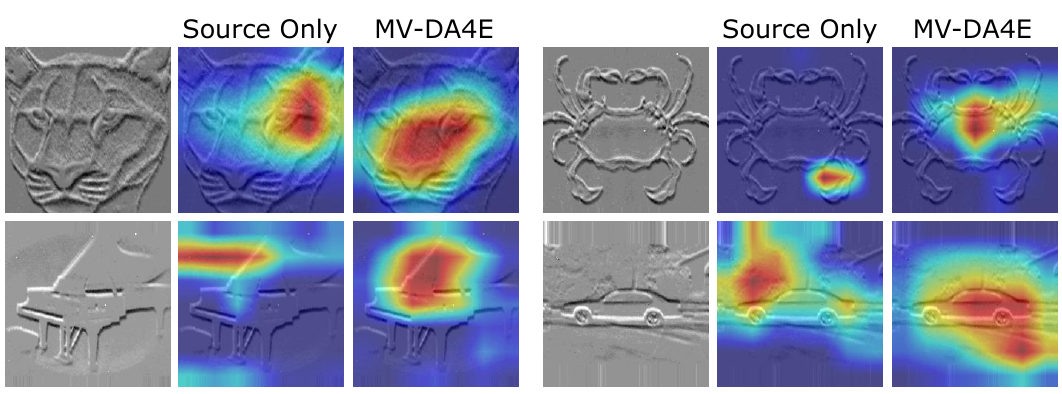}
    \caption{\new{Grad-CAM \cite{selvaraju2017grad} visualizations on several real N-Caltech101 samples. 
    In each triplet we show the input event representations (voxel grid \cite{events-zhu2019unsupervised}), the activation maps when the network is trained on simulated data only, and those obtained by training with MV-DA4E. }
    }
    \label{fig:cams}
\end{figure}
\begin{figure}
\centering
\subfloat[Source-only]{\begin{minipage}{.47\columnwidth}
    \centering
    \resizebox{\textwidth}{!}{%
    \includegraphics[width=\textwidth]{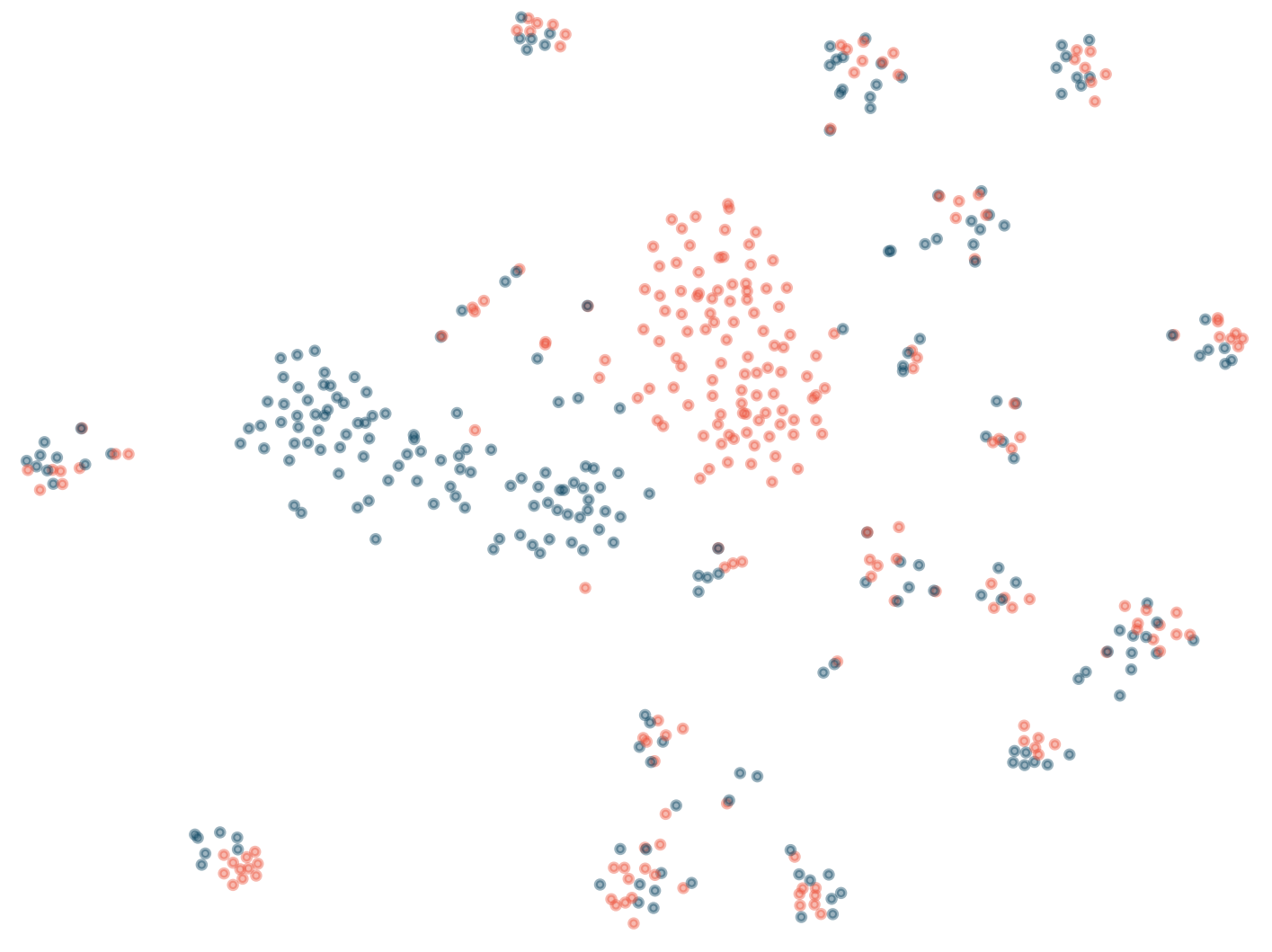}
    }
\end{minipage}}
\rule[-35px]{0.1px}{70px}
\subfloat[DA4E]{\begin{minipage}{.47\columnwidth}
    \centering
    \resizebox{\textwidth} {!} {%
    \includegraphics[width=\textwidth]{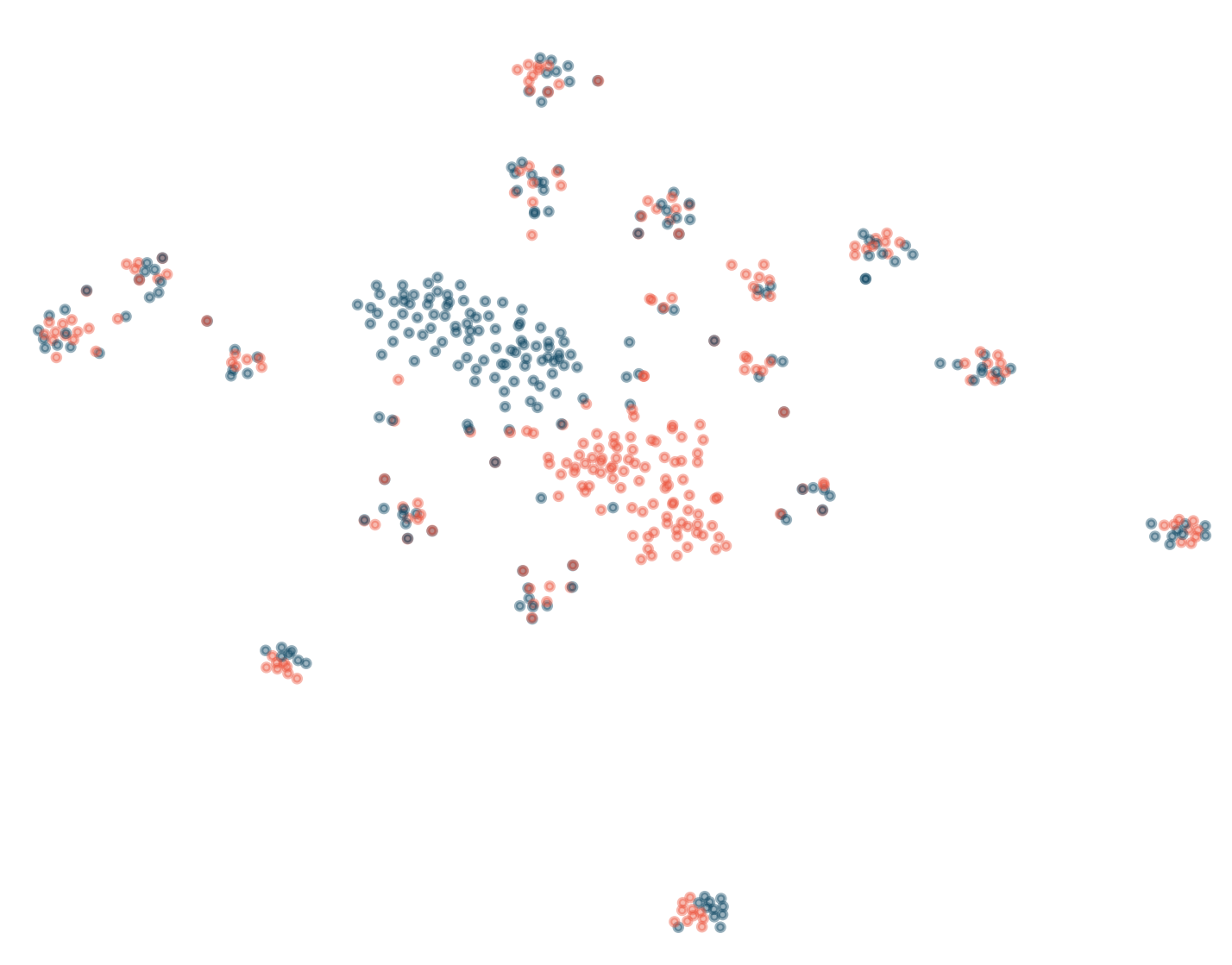}
    }
\end{minipage}}

\caption{t-SNE visualization of N-Caltech \cite{DBLP:journals/corr/OrchardJCT15} features from the last hidden layer of the main classifier. Red dots: source samples; blue dots: target samples. When adapting the two domains with the proposed DA4E (b), the two distributions align much better compared to the non-adapted case (a).}
\label{img:fig:tSNE_source_vs_da2e}
\end{figure}

Qualitative results are shown in Figure \ref{img:fig:tSNE_source_vs_da2e}, where we provide a t-SNE visualization of the source and target samples when adapting the two domains or not adapting them. \new{We also computed the Gradient-weighted Class Activation Mapping (Grad-CAM \cite{selvaraju2017grad}) on several N-Caltech101 samples, which visualize regions in the input event representation on which the network focuses the most for prediction. As shown in Figure \ref{fig:cams}, when trained with the proposed MV-DA4E approach, these regions are the most discriminative for classifying the object. }



\textit{Is the proposed multi-view approach \textit{MV-DA4E} effective?}
Table \ref{tab:da_ncaltech} shows that applying the multi-view approach \textit{MV-DA4E} significantly improves over the \textit{DA4E} configuration in all experiments, regardless of the representations and DA strategies used. These results prove the validity of the proposed method, confirming the claims made in Section~\ref{sec:mvda4e}. Interestingly, \textit{MV-DA4E} not only provides an improvement in the cross-domain scenario (Sim-to-Real), but also in the intra-domain (Supervised) one. Thus, we believe that this multi-view approach could be used as a general way to handle event representations, regardless of the task at hand.

\textit{How well our approach perform w.r.t. approaches acting on the contrast threshold C?}

\setlength{\tabcolsep}{6.5pt}
\begin{table}[t]
\begin{center}
\vspace{3mm}
\caption{\new{Target Top-1 Test Accuracy} (\%) of UDA methods w.r.t. to methods that act on the contrast threshold C.}
\label{tab:generation}
\begin{tabular}{lccccc}
\toprule\noalign{\smallskip}
\multicolumn{5}{c}{\normalsize\textsc{N-Caltech101}} \\
\noalign{\smallskip}
\cline{1-5}
\noalign{\smallskip}
Baselines & &C=$0.06$ & C=$0.15$ \cite{stoffregen2020reducing} &  C$\sim\mathcal{U}$ \cite{gehrig2020video} \\
\noalign{\smallskip}
\toprule\noalign{\smallskip}
\multirow{2}{*}{\centering Source only} 
& \textit{baseline} & 76.81 & 80.99 & 82.29 \\
& \textit{MV-baseline} & 83.12 & 84.59 & 84.93 \\
 \hline
 \hline
 \noalign{\smallskip}\noalign{\smallskip}

Our approach &w/ C values:&C=$0.06$ & C=$0.15$ & C$\sim\mathcal{U}$ \\

 \noalign{\smallskip}
\toprule\noalign{\smallskip}
\multirow{2}{*}{\centering GRL \cite{grl-pmlr-v37-ganin15}} 
&DA4E & 80.89 & 83.08 & 81.91 \\
&MV-DA4E & 84.93 & 86.77 & 86.45 \\
 \hline
 \noalign{\smallskip}
\multirow{2}{*}{\centering MMD \cite{da-mmdlong2015learning}}
&DA4E & 83.84 &86.37 &84.38 \\
&MV-DA4E & 86.94 &88.23 &87.31 \\
 \hline
 \noalign{\smallskip}
\multirow{2}{*}{\centering ROT \cite{Xu_2019}} 
&DA4E & 80.05 &79.13 & 80.36 \\
&MV-DA4E & 86.31 &86.63 &87.08\\

 \hline
 \noalign{\smallskip}
\multirow{2}{*}{\centering AFN \cite{da-afnxu2019larger}} 
&DA4E & 84.38 & 84.49 & 84.3 \\
&MV-DA4E & 87.71 &88.3 &88.17\\
 \hline
 \noalign{\smallskip}
\multirow{2}{*}{\centering Entropy \cite{grandvalet2005semi}} 
&DA4E & 85.26 & 87.0 & 85.16 \\
&MV-DA4E & \textbf{88.38} &\textbf{89.24} & \textbf{88.61}\\

\bottomrule
\end{tabular}
\end{center}
\end{table}
\setlength{\tabcolsep}{1.4pt}

{Several methods in the literature, such as \cite{gehrig2020video,stoffregen2020reducing}, address the \textit{Sim-to-Real} problem by exclusively acting on the value of the threshold $C$ used by the simulator for generating data. Since we operate with a fixed threshold, a possible question is whether our results merely derive from an optimal selection of $C$ or they stem from our choice to favor adaptation by working at the feature level. 
To answer this question, we run the different UDA approaches using voxel grid as representation and three choices for $C$, namely $C=0.06$ (the starting value used to analyze the domain shift in \cite{gehrig2020video}), $C=0.15$ (estimated following \cite{stoffregen2020reducing}), and $C\sim\mathcal{U}(0.05, 0.5)$ (as proposed in \cite{gehrig2020video}). The baselines are the $C$-only based methods (where, in particular, $C=0.15$ reproduces the settings of \cite{stoffregen2020reducing} and $C\sim\mathcal{U}$ that of \cite{gehrig2020video}).
Results in Table \ref{tab:generation} shows that (i) our approach consistently and largely outperforms the baselines for every choice of $C$, highlighting the effectiveness of approaching DA at the feature level, (ii) multi-view approaches benefits from UDA techniques in all cases and (iii) also the $C$-only based methods benefit from a multi-view approach since it contributes to significantly reduce their sensitivity to the variations of $C$.}

\input{Plots/comment_8}

\textbf{SynROD $\rightarrow$ RealROD.} In robotics, DA is used to take advantages of automatically generated synthetic data that come with ``free" annotations, to make effective prediction on real data and compensate the lack of large-scale datasets. In fact, the RGB modality tends to encode texture and appearance information, which are the characteristics that suffer most from the domain shift, and adaptation strategies are necessary to alleviate this problem. In fact, a recent line of research brought to light that ``ImageNet-trained CNNs are biased towards texture; increasing shape bias improves accuracy and  robustness"~\cite{geirhos2018imagenet}. With this in mind, we believe that the event modality could be more invariant to domain shifts, as it encodes additional geometric and temporal information, and it is more robust to lighting and color variations. Moreover, it has been demonstrated that exploiting the complementarity of a multi-modal input leads to better adaptation performance in this cross-domain scenario \cite{loghmani2020unsupervised}. Thus, in order to verify the soundness of event data extracted from RGB images, and their potentiality in real-world applications, we analyze the behaviour of event modality (both single and multi-modal RGB+Event) in the SynROD $\rightarrow$ RealROD scenario. In order to assess the benefit of the event modality, we compare it to the traditional ones, i.e., RGB and depth. 

For these analysis, we chose the representation which revealed better performances across domains, i.e., VoxelGrid, and the multi-view approach \textit{MV-DA4E}, which shown its superiority in all the experiments performed on event data (Table \ref{tab:da_ncaltech}). 
From the results shown in Table \ref{tab:da_rod}, it is evident that the event modality is more robust than depth modality, and we also remark that UDA techniques work well in this setting too. In particular, we show that the event modality is less sensitive to domain changes w.r.t. the depth modality, both in the single modal ($7.56\%$ \textit{vs} $39.43\%$) and multi-modal RGB+E ($47.7\%$ \textit{vs} $52.87\%$) scenarios, when no DA techniques are applied. Interestingly, when combined to the RGB modality, event data slightly improves the model accuracy, differently from depth modality which causes a degradation of performance as the network find difficulties in exploiting the complementarity of the two. Furthermore, we highlight that on average UDA performance on RGB-E is greater than the one obtained with RGB-D.

\setlength{\tabcolsep}{7pt}
\begin{table}[t]
\begin{center}
\vspace{3mm}
\caption{\new{Target Top-1 Accuracy} (\%) of the event, RGB and depth modalities, both in single-modal and multi-modal (RGB+E).}
\label{tab:da_rod}
\begin{tabular}{llcccc}
\toprule\noalign{\smallskip}
\multicolumn{6}{c}{\normalsize\textsc{SynROD $\implies$ ROD}} \\
\noalign{\smallskip}
\cline{1-6}
\noalign{\smallskip}
Method & RGB & Depth & {Event} &{RGB+E} & {RGB+D} \\
\noalign{\smallskip}
\toprule\noalign{\smallskip}
\multirow{1}{*}{\centering Source only} 
 & 52.13 & 7.56 & 39.43  & 52.87 & 47.7\\
 
 \hline
 \noalign{\smallskip}
\multirow{1}{*}{\centering GRL \cite{grl-pmlr-v37-ganin15}} 
 & 57.12 & 26.11 & 46.15 & 55.11 & 59.51 \\

 \hline
 \noalign{\smallskip}
\multirow{1}{*}{\centering MMD \cite{da-mmdlong2015learning}}
 & 63.68 & 29.34 & 47.52 & 62.39 & 62.57 \\

 \hline
 \noalign{\smallskip}
\multirow{1}{*}{\centering Rotation \cite{Xu_2019}\cite{loghmani2020unsupervised}} 
 & 63.21 & 6.70 & 41.84  & 66.68 & 66.68  \\
 
 \hline
 \noalign{\smallskip}
\multirow{1}{*}{\centering AFN \cite{da-afnxu2019larger}} 
 & 64.63 & 30.72 & 52.38 & 66.87 & 62.4    \\
  \hline
 \noalign{\smallskip}
\multirow{1}{*}{\centering Entropy \cite{grandvalet2005semi}} 
 & 61.53& 16.79 & 49.23& 66.23 &   63.12  \\
 \hline
 \noalign{\smallskip}
\multirow{1}{*}{\centering Avg } 
 & 62.03 & 21.93 & 47.42 & \textbf{63.46} & 62.86   \\

\bottomrule
\end{tabular}
\end{center}
\end{table}
\setlength{\tabcolsep}{1.4pt}

%% file: Plots/comment_8.tex
\begin{figure}[t!]
\centering
\begin{minipage}[t]{0.23\textwidth}
     \resizebox {\columnwidth} {!} {
        \begin{tikzpicture}
        \pgfplotsset{every axis legend/.append style={
at={(0.5,1.03)},
anchor=south}}
        \begin{axis}[
          enlargelimits=false,
          ylabel={ Top-1 Accuracy (\%)},
          xlabel={Target data ($\%$)},
           xmin=-7, xmax=107,
           ymin=70, ymax=94,
          xtick={0,25,50,75,100},
          ytick={70,80,90},
          ymajorgrids=true,
          grid style=dashed,
          width=6.5cm,
          height=4cm,
          legend columns=-1,
                     legend style={draw=none},
                     every axis plot/.append style={ultra thick}
        ]
        \addplot[
          color=MaterialBlue800,
          mark=square*]
        table[x index=0,y index=1,col sep=comma]
        {Plots/ent.txt};
        \addplot[dashed][
          color=MaterialOrange500,
          mark=oplus*]
        table[x index=0,y index=1,col sep=comma]
        {Plots/source_only.txt};
        \legend{DA4E,SourceOnly} 
        \end{axis}
        \end{tikzpicture}%
      }
    \end{minipage}
    \begin{minipage}[t]{0.23\textwidth}
     \resizebox {\columnwidth} {!} {
        \begin{tikzpicture}
        \pgfplotsset{every axis legend/.append style={
at={(0.5,1.03)},
anchor=south}}
        \begin{axis}[
          enlargelimits=false,
          ylabel={ Top-1 Accuracy (\%)},
          xlabel={Target data ($\%$)},
           xmin=-7, xmax=107,
           ymin=80, ymax=92,
          xtick={0,25,50,75,100},
          ytick={80,85,90},
          ymajorgrids=true,
          grid style=dashed,
          width=6.5cm,
          height=4cm,
          legend columns=-1,
           legend style={draw=none},
           every axis plot/.append style={ultra thick}
        ]
        \addplot[
          color=MaterialBlue800,
          mark=square*]
        table[x index=0,y index=1,col sep=comma]
        {Plots/ent_mv.txt};
        \addplot[dashed][
          color=MaterialOrange500,
          mark=oplus*]
        table[x index=0,y index=1,col sep=comma]
        {Plots/source_only_mv.txt};
        \legend{MV-DA4E,SourceOnly}

        \end{axis}

        \end{tikzpicture}%
      }
    \label{fig:lambda_var}
    \end{minipage}
    
\vspace{-0.5cm}
\caption{\new{Difference in terms of performance based on percentage ($\%$) of target data used during training, obtained with constant threshold $C=0.06$.} }
\label{fig:comment_8}
\end{figure}

%% file: Sections/5.Conclusions.tex
\section{Conclusions}
\label{sec:conclusions}
In this work we propose an alternative way of answering a very recent research problem regarding how to bridge Sim-to-Real gap for event cameras arising from event generation. By seeing the problem under a new prospective, the domain shift, we show that Unsupervised Domain Adaptation (UDA) techniques working at feature level are an effective way of tackling this issue, w.r.t. previous work that act on the input level. Moreover we propose a multi-view approach to deal with event representations, which outperforms existing methods and proved to work well in conjunction with other UDA strategies. We validate both approaches through extensive experiments on the N-Caltech101 dataset and the popular RGB-D Object Dataset (ROD). We hope that our work will ignite future works on the opportunity of using UDA methods for event data.


%% file: root.bbl
\begin{thebibliography}{10}
\providecommand{\url}[1]{#1}
\csname url@rmstyle\endcsname
\providecommand{\newblock}{\relax}
\providecommand{\bibinfo}[2]{#2}
\providecommand\BIBentrySTDinterwordspacing{\spaceskip=0pt\relax}
\providecommand\BIBentryALTinterwordstretchfactor{4}
\providecommand\BIBentryALTinterwordspacing{\spaceskip=\fontdimen2\font plus
\BIBentryALTinterwordstretchfactor\fontdimen3\font minus
  \fontdimen4\font\relax}
\providecommand\BIBforeignlanguage[2]{{%
\expandafter\ifx\csname l@#1\endcsname\relax
\typeout{** WARNING: IEEEtran.bst: No hyphenation pattern has been}%
\typeout{** loaded for the language `#1'. Using the pattern for}%
\typeout{** the default language instead.}%
\else
\language=\csname l@#1\endcsname
\fi
#2}}

\bibitem{maqueda2018event}
A.~I. Maqueda, A.~Loquercio, G.~Gallego, N.~Garc{\'\i}a, and D.~Scaramuzza,
  ``Event-based vision meets deep learning on steering prediction for
  self-driving cars,'' in \emph{Proceedings of the IEEE Conference on Computer
  Vision and Pattern Recognition}, 2018, pp. 5419--5427.

\bibitem{gehrig2019end}
D.~Gehrig, A.~Loquercio, K.~G. Derpanis, and D.~Scaramuzza, ``End-to-end
  learning of representations for asynchronous event-based data,'' in
  \emph{Proceedings of the IEEE/CVF International Conference on Computer
  Vision}, 2019, pp. 5633--5643.

\bibitem{rebecq2018esim}
H.~Rebecq, D.~Gehrig, and D.~Scaramuzza, ``Esim: an open event camera
  simulator,'' in \emph{Conference on Robot Learning}.\hskip 1em plus 0.5em
  minus 0.4em\relax PMLR, 2018, pp. 969--982.

\bibitem{gehrig2020video}
D.~Gehrig, M.~Gehrig, J.~Hidalgo-Carri{\'o}, and D.~Scaramuzza, ``Video to
  events: Recycling video datasets for event cameras,'' in \emph{Proceedings of
  the IEEE/CVF Conference on Computer Vision and Pattern Recognition}, 2020,
  pp. 3586--3595.

\bibitem{stoffregen2020reducing}
T.~Stoffregen, C.~Scheerlinck, D.~Scaramuzza, T.~Drummond, N.~Barnes,
  L.~Kleeman, and R.~Mahony, ``Reducing the sim-to-real gap for event
  cameras,'' in \emph{Proc. Eur. Conf. Comput. Vis.}\hskip 1em plus 0.5em minus
  0.4em\relax Springer, 2020, pp. 534--549.

\bibitem{grl-pmlr-v37-ganin15}
Y.~Ganin and V.~Lempitsky, ``Unsupervised domain adaptation by
  backpropagation,'' ser. Proceedings of Machine Learning Research, F.~Bach and
  D.~Blei, Eds., vol.~37.\hskip 1em plus 0.5em minus 0.4em\relax Lille, France:
  PMLR, 07--09 Jul 2015, pp. 1180--1189.

\bibitem{da-mmdlong2015learning}
M.~Long, Y.~Cao, J.~Wang, and M.~Jordan, ``Learning transferable features with
  deep adaptation networks,'' in \emph{International conference on machine
  learning}.\hskip 1em plus 0.5em minus 0.4em\relax PMLR, 2015, pp. 97--105.

\bibitem{loghmani2020unsupervised}
M.~R. Loghmani, L.~Robbiano, M.~Planamente, K.~Park, B.~Caputo, and M.~Vincze,
  ``Unsupervised domain adaptation through inter-modal rotation for rgb-d
  object recognition,'' \emph{IEEE Robotics and Automation Letters}, vol.~5,
  no.~4, pp. 6631--6638, 2020.

\bibitem{da-afnxu2019larger}
R.~Xu, G.~Li, J.~Yang, and L.~Lin, ``Larger norm more transferable: An adaptive
  feature norm approach for unsupervised domain adaptation,'' in
  \emph{Proceedings of the IEEE International Conference on Computer Vision},
  2019, pp. 1426--1435.

\bibitem{grandvalet2005semi}
Y.~Grandvalet and Y.~Bengio, ``Semi-supervised learning by entropy
  minimization,'' in \emph{Adv. Neural Inform. Process. Syst.}, vol. 367, 01
  2004, pp. 281--296.

\bibitem{DBLP:journals/corr/OrchardJCT15}
G.~Orchard, A.~Jayawant, G.~K. Cohen, and N.~Thakor, ``Converting static image
  datasets to spiking neuromorphic datasets using saccades,'' \emph{Frontiers
  in neuroscience}, vol.~9, p. 437, 2015.

\bibitem{lai2011large}
K.~Lai, L.~Bo, X.~Ren, and D.~Fox, ``A large-scale hierarchical multi-view
  rgb-d object dataset,'' in \emph{2011 IEEE international conference on
  robotics and automation}.\hskip 1em plus 0.5em minus 0.4em\relax IEEE, 2011,
  pp. 1817--1824.

\bibitem{da-mcdsaito2018maximum}
K.~Saito, K.~Watanabe, Y.~Ushiku, and T.~Harada, ``Maximum classifier
  discrepancy for unsupervised domain adaptation,'' in \emph{Proceedings of the
  IEEE Conference on Computer Vision and Pattern Recognition}, 2018, pp.
  3723--3732.

\bibitem{da-adv-deng2019cluster}
Z.~Deng, Y.~Luo, and J.~Zhu, ``Cluster alignment with a teacher for
  unsupervised domain adaptation,'' in \emph{Proceedings of the IEEE
  International Conference on Computer Vision}, 2019, pp. 9944--9953.

\bibitem{da-adv-tang2020discriminative}
H.~Tang and K.~Jia, ``Discriminative adversarial domain adaptation.'' in
  \emph{AAAI}, 2020, pp. 5940--5947.

\bibitem{Xu_2019}
X.~Jiaolong, X.~Liang, and A.~M. López, ``Self-supervised domain adaptation
  for computer vision tasks,'' \emph{IEEE Access}, vol.~7, pp.
  156\,694--156\,706, 2019.

\bibitem{carlucci2019domain}
F.~M. Carlucci, A.~D'Innocente, S.~Bucci, B.~Caputo, and T.~Tommasi, ``Domain
  generalization by solving jigsaw puzzles,'' in \emph{Proceedings of the IEEE
  Conference on Computer Vision and Pattern Recognition}, 2019, pp. 2229--2238.

\bibitem{DBLP:journals/corr/BousmalisTSKE16}
K.~Bousmalis, G.~Trigeorgis, N.~Silberman, D.~Krishnan, and D.~Erhan, ``Domain
  separation networks,'' \emph{Advances in neural information processing
  systems}, vol.~29, pp. 343--351, 2016.

\bibitem{blender2018}
\BIBentryALTinterwordspacing
``Blender,'' {Accessed: Feb. 24, 2021}. [Online]. Available:
  \url{https://www.blender.org}
\BIBentrySTDinterwordspacing

\bibitem{ros2016synthia}
G.~Ros, L.~Sellart, J.~Materzynska, D.~Vazquez, and A.~M. Lopez, ``The synthia
  dataset: A large collection of synthetic images for semantic segmentation of
  urban scenes,'' in \emph{Proceedings of the IEEE conference on computer
  vision and pattern recognition}, 2016, pp. 3234--3243.

\bibitem{richter2016playing}
S.~R. Richter, V.~Vineet, S.~Roth, and V.~Koltun, ``Playing for data: Ground
  truth from computer games,'' in \emph{European conference on computer
  vision}.\hskip 1em plus 0.5em minus 0.4em\relax Springer, 2016, pp. 102--118.

\bibitem{alberti2020idda}
E.~Alberti, A.~Tavera, C.~Masone, and B.~Caputo, ``Idda: a large-scale
  multi-domain dataset for autonomous driving,'' \emph{IEEE Robotics and
  Automation Letters}, vol.~5, no.~4, pp. 5526--5533, 2020.

\bibitem{cordts2016cityscapes}
M.~Cordts, M.~Omran, S.~Ramos, T.~Rehfeld, M.~Enzweiler, R.~Benenson,
  U.~Franke, S.~Roth, and B.~Schiele, ``The cityscapes dataset for semantic
  urban scene understanding,'' in \emph{Proceedings of the IEEE conference on
  computer vision and pattern recognition}, 2016, pp. 3213--3223.

\bibitem{hoffman2016fcns}
J.~Hoffman, D.~Wang, F.~Yu, and T.~Darrell, ``Fcns in the wild: Pixel-level
  adversarial and constraint-based adaptation,'' \emph{arXiv preprint
  arXiv:1612.02649}, 2016.

\bibitem{tsai2018learning}
Y.-H. Tsai, W.-C. Hung, S.~Schulter, K.~Sohn, M.-H. Yang, and M.~Chandraker,
  ``Learning to adapt structured output space for semantic segmentation,'' in
  \emph{Proceedings of the IEEE conference on computer vision and pattern
  recognition}, 2018, pp. 7472--7481.

\bibitem{vu2019advent}
T.-H. Vu, H.~Jain, M.~Bucher, M.~Cord, and P.~P{\'e}rez, ``Advent: Adversarial
  entropy minimization for domain adaptation in semantic segmentation,'' in
  \emph{Proceedings of the IEEE/CVF Conference on Computer Vision and Pattern
  Recognition}, 2019, pp. 2517--2526.

\bibitem{hoffman2018cycada}
J.~Hoffman, E.~Tzeng, T.~Park, J.-Y. Zhu, P.~Isola, K.~Saenko, A.~Efros, and
  T.~Darrell, ``Cycada: Cycle-consistent adversarial domain adaptation,'' in
  \emph{International conference on machine learning}.\hskip 1em plus 0.5em
  minus 0.4em\relax PMLR, 2018, pp. 1989--1998.

\bibitem{DBLP:journals/corr/abs-1903-01212}
J.~Wang and K.~Zhang, ``Unsupervised domain adaptation learning algorithm for
  rgb-d stairway recognition,'' \emph{Instrumentation}, vol. 6(2), pp. 21--29,
  2019.

\bibitem{10.1016/j.sigpro.2016.07.018}
X.~Li, M.~Fang, J.-J. Zhang, and J.~Wu, ``Domain adaptation from rgb-d to rgb
  images,'' \emph{Signal Process.}, vol. 131, no.~C, p. 27–35, Feb. 2017.

\bibitem{events-cohen2016thesis}
G.~K. Cohen, ``{Event-Based Feature Detection, Recognition and
  Classification},'' Theses, {Universit{\'e} Pierre et Marie Curie - Paris VI ;
  University of Western Sydney}, Sept. 2016.

\bibitem{events-cannici2019asynchronous}
M.~Cannici, M.~Ciccone, A.~Romanoni, and M.~Matteucci, ``Asynchronous
  convolutional networks for object detection in neuromorphic cameras,'' in
  \emph{Proceedings of the IEEE/CVF Conference on Computer Vision and Pattern
  Recognition Workshops}, 2019, pp. 0--0.

\bibitem{events-lagorce2016hots}
X.~Lagorce, G.~Orchard, F.~Galluppi, B.~E. Shi, and R.~B. Benosman, ``Hots: a
  hierarchy of event-based time-surfaces for pattern recognition,'' \emph{IEEE
  transactions on pattern analysis and machine intelligence}, vol.~39, no.~7,
  pp. 1346--1359, 2016.

\bibitem{events-sironi2018hats}
A.~Sironi, M.~Brambilla, N.~Bourdis, X.~Lagorce, and R.~Benosman, ``Hats:
  Histograms of averaged time surfaces for robust event-based object
  classification,'' in \emph{Proceedings of the IEEE Conference on Computer
  Vision and Pattern Recognition}, 2018, pp. 1731--1740.

\bibitem{events-zhu2019unsupervised}
A.~Z. Zhu, L.~Yuan, K.~Chaney, and K.~Daniilidis, ``Unsupervised event-based
  learning of optical flow, depth, and egomotion,'' in \emph{Proceedings of the
  IEEE/CVF Conference on Computer Vision and Pattern Recognition}, 2019, pp.
  989--997.

\bibitem{events-innocenti2020temporal}
S.~U. Innocenti, F.~Becattini, F.~Pernici, and A.~D. Bimbo, ``Temporal binary
  representation for event-based action recognition,'' \emph{arXiv}, 2020.

\bibitem{events-gehrig2019end}
D.~Gehrig, A.~Loquercio, K.~G. Derpanis, and D.~Scaramuzza, ``End-to-end
  learning of representations for asynchronous event-based data,'' in
  \emph{Proceedings of the IEEE/CVF International Conference on Computer
  Vision}, 2019, pp. 5633--5643.

\bibitem{events-cannici2020differentiable}
M.~Cannici, M.~Ciccone, A.~Romanoni, and M.~Matteucci, ``A differentiable
  recurrent surface for asynchronous event-based data,'' in \emph{European
  Conference on Computer Vision}.\hskip 1em plus 0.5em minus 0.4em\relax
  Springer, 2020, pp. 136--152.

\bibitem{events-deng2020amae}
Y.~Deng, Y.~Li, and H.~Chen, ``Amae: Adaptive motion-agnostic encoder for
  event-based object classification,'' \emph{IEEE Robotics and Automation
  Letters}, vol.~5, no.~3, pp. 4596--4603, 2020.

\bibitem{rebecq2019events}
H.~Rebecq, R.~Ranftl, V.~Koltun, and D.~Scaramuzza, ``Events-to-video: Bringing
  modern computer vision to event cameras,'' in \emph{Proceedings of the
  IEEE/CVF Conference on Computer Vision and Pattern Recognition}, 2019, pp.
  3857--3866.

\bibitem{DBLP:journals/corr/abs-1906-07165}
------, ``High speed and high dynamic range video with an event camera,''
  \emph{{IEEE} Trans. Pattern Anal. Mach. Intell. (T-PAMI)}, 2019.

\bibitem{cannici2021nrod}
M.~Cannici, C.~Plizzari, M.~Planamente, M.~Ciccone, A.~Bottino, B.~Caputo, and
  M.~Matteucci, ``N-rod: a neuromorphic dataset for synthetic-to-real domain
  adaptation,'' in \emph{Proceedings of the IEEE/CVF Conference on Computer
  Vision and Pattern Recognition Workshops}, 2021, pp. 0--0.

\bibitem{yosinski2014transferable}
J.~Yosinski, J.~Clune, Y.~Bengio, and H.~Lipson, ``How transferable are
  features in deep neural networks?'' \emph{arXiv preprint arXiv:1411.1792},
  2014.

\bibitem{fei2006one}
L.~Fei-Fei, R.~Fergus, and P.~Perona, ``One-shot learning of object
  categories,'' \emph{IEEE transactions on pattern analysis and machine
  intelligence}, vol.~28, no.~4, pp. 594--611, 2006.

\bibitem{hochreiter1997long}
S.~Hochreiter and J.~Schmidhuber, ``Long short-term memory,'' \emph{Neural
  computation}, vol.~9, no.~8, pp. 1735--1780, 1997.

\bibitem{he2016deep}
K.~He, X.~Zhang, S.~Ren, and J.~Sun, ``Deep residual learning for image
  recognition,'' in \emph{Proceedings of the IEEE conference on computer vision
  and pattern recognition}, 2016, pp. 770--778.

\bibitem{noroozi2016unsupervised}
M.~Noroozi and P.~Favaro, ``Unsupervised learning of visual representations by
  solving jigsaw puzzles,'' in \emph{European conference on computer
  vision}.\hskip 1em plus 0.5em minus 0.4em\relax Springer, 2016, pp. 69--84.

\bibitem{selvaraju2017grad}
R.~R. Selvaraju, M.~Cogswell, A.~Das, R.~Vedantam, D.~Parikh, and D.~Batra,
  ``Grad-cam: Visual explanations from deep networks via gradient-based
  localization,'' in \emph{Proceedings of the IEEE international conference on
  computer vision}, 2017, pp. 618--626.

\bibitem{geirhos2018imagenet}
R.~Geirhos, P.~Rubisch, C.~Michaelis, M.~Bethge, F.~A. Wichmann, and
  W.~Brendel, ``Imagenet-trained cnns are biased towards texture; increasing
  shape bias improves accuracy and robustness,'' \emph{arXiv preprint
  arXiv:1811.12231}, 2018.

\end{thebibliography}
